# The experience of humans' and robots' mutual (im)politeness in enacted service scenarios: An empirical study

Victor Kaptelinin[0000-0002-5326-7054], Department of Informatics, Umeå University, Umeå, Sweden (contact author, email: victor.kaptelinin@umu.se)

Suna Bensch, Department of Computing Science, Umeå University, Umeå, Sweden

Thomas Hellström, Department of Computing Science, Umeå University, Umeå, Sweden

Patrik Björnfot, Department of Informatics, Umeå University, Umeå, Sweden

Shikhar Kumar, Department of Industrial Engineering and Management, Ben-Gurion University of the Negev, Israel

## ABSTRACT

The paper reports an empirical study of the effect of human treatment of a robot on the social perception of the robot's behavior. The study employed an enacted interaction between an anthropomorphic "waiter" robot and two customers. The robot and one of the customers (acted out by a researcher) were following four different interaction scripts, representing all combinations of mutual politeness and impoliteness of the robot and the customer. The participants (N=24, within-subject design) were assigned the role of an "included observer", that is, a fellow customer who was present in the situation without being actively involved in the interactions. The participants assessed how they experienced the interaction scenarios by providing Likert scale scores and free-text responses. The results indicate that while impolite robots' behavior was generally assessed negatively, it was commonly perceived as more justifiable and fairer if the robot was treated impolitely by the human. Politeness reciprocity expectations in the context of the social perception of robots are discussed.

## 1. Introduction

The view of intelligent agents as social actors, which are expected to follow certain social rules and should possess some kind of social competence, has gained significant ground in Human-Robot Interaction (HRI) and related fields (e.g., [9,28]). A wide range of conceptual analyses, empirical studies, and design activities have been driven by the vision of intelligent agents, capable of understanding the social context at hand and acting appropriately.

Arguably, agents' social competence cannot be limited to an ability to adapt to the given, pre-existing context of human-human interactions. The very fact that an agent enters a setting as a social actor brings in a range of new social phenomena. In particular, people commonly manifest social behavior directed at an intelligent agent, e.g., treat it as a communication or collaboration partner (e.g., [8,34]). Moreover, an agent's interaction with a human may be perceived by other people, and thus make an impact on the social context of the setting beyond that particular interaction [27].

A relevant theme for HRI research, therefore, is how a socially competent intelligent agent should respond to *social behavior and expectations, directed at the agent itself.* This paper contributes to research in this direction by focusing on *politeness reciprocity* in robots, that is, on whether a robot's behavior toward a human should be expected to correspond to the politeness (or lack thereof) of the human. With some notable exceptions (e.g., [10]), the issue received limited attention in existing HRI research, as there has been little overlap between research on "politeness" and "reciprocity". On the one hand, while numerous HRI studies have investigated robots' perceived politeness and optimal politeness strategies, the research typically does not deal with reciprocity. On the other hand, while studies in HCI [28] and HRI [36,37] demonstrated that reciprocity is inherent to not only human-human, but also human-technology interaction, the research has been mostly concerned with the exchange of favors between rational actors rather than politeness as such. To the best of our knowledge, there have been no previous systematic empirical analyses of robots' politeness reciprocity in HRI research.

The study reported in this paper addresses this gap in existing research by analyzing the effect of the human treatment of a social robot on the perception of the robot's behavior. The study specifically focused on the experience of people, who mostly *observed another person interacting with a robot* rather than actively interacted with the robot themselves. The main rationale for choosing this approach was that in real-life contexts, people who observe (im)polite interactions are commonly affected by them and the impact of such interactions on the observers (or its anticipation) can be a major factor affecting the interactions per se. Therefore, understanding the experience of people perceiving human-robot interactions "from the outside" can be considered a valid object of study in HRI. From a methodological point of view, an advantage of the strategy was that it made it possible for us to expose different participants to the same scenarios of human-robot interaction. As discussed below, adopting this approach also helped us deal with ethical research challenges, associated with studying impolite human-robot interactions.

The study was designed as a mixed-method experiment, in which different levels of human politeness were systematically combined with different levels of robot politeness. The following hypotheses were tested.

H1. *Social attributes of a robot are perceived differently depending on whether the robot is treated politely or impolitely*. The hypothesis was tested by using quantitative Likert-scale data for the conditions of the study to determine whether there was a statistically significant interaction between the factors of "human politeness" and "robot politeness".

H2: *Human assessments of a robot's behavior are affected by an expectation of politeness reciprocity*. The hypothesis was tested by using qualitative free-response data to find out whether a robot's behavior was assessed by the participants as more natural (e.g., "understandable" or "justifiable") when the robot treated a human the way it was treated by the human.

## 2. Theoretical foundations and related research

This section provides an overview of relevant existing work, including (a) approaches in social sciences, adopted in HRI to help operationalize the notion of "politeness", (b) HRI research on politeness, including implementation strategies, as well as effects, of robot politeness, and (c) analyses of reciprocity in HRI research.

## 2.1. Approaches to politeness, adopted in HRI

Several theoretical approaches to politeness, developed in social sciences, were adopted in HRI to go beyond the general notion of "being nice" and define politeness in a more rigorous and differentiated way. The most influential approaches have been the ones proposed by Grice [2], Lakoff [24], and Brown and Levinson [6][1].

Grice's Cooperative Principle [2] states that certain rules, or "maxims", should be followed to ensure efficient communication. Grice proposes four maxims, to which all communication parties should adhere: Quantity (provide enough information but not too much), Quality (be truthful), Relation (be relevant), and Manner (be clear and brief).

Gricean maxims essentially stipulate that communication is to be conducted in good faith. Therefore, following them can be considered a way of being nice to other people. However, the maxims do not offer an explicit and general conceptual account of politeness as such. A more comprehensive approach, specifically dealing with politeness, was proposed by Lakoff [24], who identified "being polite" as a separate aspect of conversation, different from "being clear". According to Lakoff, politeness can be defined as adhering to the following three rules: "Don't impose", "Give options", and "Be friendly". The rules, basically, define politeness as not telling people how to do their business, providing them with a choice, and communicating the feelings of closeness and solidarity to make the communication partners feel good.

Politeness theory by Brown and Levinson [6] offers an approach to politeness, based on an advanced conceptual framework describing the origins and mechanisms of polite behavior. The theory employs Goffman's notion of "face" [6] that refers to the positive self-image that a person maintains in social interactions. A fundamental distinction is made between *negative face*, the desire to be unimpeded in one's actions, and *positive face*, the desire to be approved of. The theory defines minimizing threats to negative and positive face (or "face threatening acts") as, respectively, *negative politeness* and *positive politeness*. The conceptual framework is employed by Brown and Levinson to identify specific types and concrete instances of politeness strategies.

Each of these approaches have informed HRI research, mostly by providing guidelines for implementing polite robots' behavior. As discussed in the next section, different studies adopted different approaches (or their combinations), with Brown and Levinson's politeness theory [6] being the most popular theoretical foundation for studies of politeness in HRI. The theory is also a key theoretical influence for the study, reported in this paper. A significant advantage of the theory in the context of the study is that its concepts of "face" and "face threatening acts" provide a valuable support for implementing both polite and impolite robot's behavior.

Recently, Bar-Or et al. [2] questioned the suitability of politeness theory in human-computer interaction, in particular, on the grounds that the concept of "face", integral to the theory, is too strong for the context of HCI. Instead, Bar-Or et al. propose a framework for studying politeness in HCI, based on a combination of Gricean maxims and Lakoff's politeness rules. In our view, while the argument may be valid for human interaction with certain types of technologies, such as websites or computer applications, the concept of "face" (and, accordingly, politeness theory) can be more at home in HRI, and especially in social robotics.

## 2.2. HRI research on politeness

HRI research on politeness can be roughly divided into studies of *human politeness*, that is, how politely people are, or should be, treating robots, and *robot politeness*, that is, how politely robots are, or should be, treating people. Studies of the first type were mostly dealing with one particular topic relevant to politeness, namely, violence against robots. It was found, for instance, that intelligent robots were perceived as being "sort of alive," which reduced the users' willingness to "kill" them [3]. Reasons for banning violence against robots were analyzed in [27].

Studies of the second type, dealing with robot politeness, comprise the majority of politeness research in HRI. The studies were mostly addressing various aspects of "socially competent" or "socially appropriate" robot behavior. Given the centrality of politeness in socially competent behavior in humans, it is not surprising that politeness has been a key theme in research dealing with socially competent intelligent agents. Key issues addressed in studies of robot politeness were (a) the implementation and assessment of robot politeness and (b) the effects of robot politeness on human perception and behavior.

### 2.2.1. Robot politeness implementation and assessment

Various strategies for designing robots and their behaviors that can be experienced as being polite were explored in HRI research. The most common approach was employing robotic speech and using different linguistic forms of spoken utterances. Examples of studies using this approach are [5,13,15,21,23,30,32,35,38,39]. Theories imported to HRI from social sciences, such as Brown and Levinson's politeness theory [6] and Lakoff's politeness principles [24], mentioned above, informed several attempts to implement robots' politeness through verbal behavior. Other ways of implementing robot politeness, reported in HRI studies, include using pointing [26], displayed text [1], body poses [29], and various motion strategies [19].

A common conclusion from numerous HRI studies was that politeness interacts with, and may be difficult to differentiate from, other aspects of robot behavior, such as understandability of the robot's actions [26], fairness [16], and the degree to which the robot expresses social cues, such as voice tone and head movements [12]. An unclear distinction between politeness, fairness, and etiquette was also pointed out in [14] and [16].

---

[1] A detailed discussion of current theoretical approaches to politeness in human-technology interaction research can be found in Bar-Or et al. [2].

Several instruments for measuring human subjective experience of robots' social attributes were developed in HRI [20], including Godspeed [4] and RoSAS [7]. The Godspeed scale, which was created to provide a standardized measurement of human perception of service robots, includes sets of semantic differential scales for each of the following key factors: anthropomorphism, animacy, likeability, perceived intelligence, and perceived safety. RoSAS (The Robotic Social Attribute Scale), aiming at measuring social perception of robots and building on Godspeed, comprises 18 nine-point Likert scales organized around 3 factors: warmth, competence, and discomfort. While Godspeed and RoSAS provide a helpful starting point for analyses of politeness reciprocity in human-robot interactions, the scales do not specifically focus on politeness and social appropriateness of robots' behavior.

#### 2.2.2. Effects of robot politeness

Studies on the effects of robot politeness were focusing on how politeness affected (a) humans' perception of the robot, (b) their behavior toward robots, and (c) humans' own tasks.

Polite robot behavior was generally found to have a positive impact on how individuals perceive, and interact with, robots. For instance, polite guards were assessed as friendlier and less intimidating [16]. Polite robots, initiating interaction with people, were doing it more successfully and were perceived as less intrusive [19]. Interactions with polite robots were found to be associated with higher levels of enjoyment and trust [22] and thus positively affecting the perceived persuasiveness of a robot. People were also more willing to give polite robots priority in an elevator [1] and let polite robot pass in crowded environments [5].

Evidence indicates that the politeness strategies employed by the robots are also important. Srinvasan and Takayama [38] found Brown and Levinson's [6] positive politeness strategy to be the most effective one in a situation where a robot asked a human for help. Cross-cultural HRI studies found a correlation between preferences for politeness strategies in human-robot and human-human interaction [21,35].

Findings regarding the effect of robot politeness on encouraging people to do their own tasks have been mixed. Lee et al. [25] showed that social robots' use of direct speech combined with polite gestures generally increased patient compliance with healthcare advice. However, studies where robots provided encouragement for the participants to do physically challenging or boring tasks did not show an advantage of polite robot behavior [15,30].

### 2.3. Reciprocity in HRI research

Reciprocity can be generally described as responding nicely and cooperatively when being treated nicely, and responding conversely in the opposite case [37]. A major aspect of human social life, reciprocity was found to be present not only in human-human interactions, but also in human interactions with technology. In the field of human-computer interaction (HCI), a 1997 study by Fogg and Nass [11] showed that the help provided by a person to a computer depended on the help that was previously received by the person from this particular computer.

More recently, reciprocity transpired as a significant research theme in HRI [17,18,36,37]. A study involving preschool children [17], found that interactions, categorized as "Attempts at reciprocity" [2], were more common when the children were playing with the robotic dog AIBO than when they were playing with a stuffed dog. A conceptual analysis by Kahn et al. [18], aiming at elucidating the psychological benchmarks by which to measure success in creating humanoid robots, highlighted reciprocity as a key such benchmark.

Study by Sandoval et al. [37] employed versions of the Prisoner's Dilemma Game and the Ultimatum Game, played by the participants with robots and human partners using different collaboration strategies. While the participants generally preferred human partners, they were equally reciprocal to humans and robots when responding to their collaboration strategies. Empirical evidence from another study that used a version of the Ultimate Game [36] suggested that there was a relationship between robots' likeability and the reciprocity they manifested in collaborative game playing.

Most HRI studies, investigating reciprocity, did not explicitly deal with politeness. Instead, they were focusing, for instance, on exchanging favors through behavior driven by rational self-interest [36,37]. A rare example of an analysis dealing with both reciprocity and politeness is a focus group study by Draper and Sorell [10], in which the participants, older adults and their caretakers, discussed the possibility of using robots to modify older adults' rude behavior. The participants were found to express mixed opinions and assessments regarding an imaginary case of a robot, which refused to comply with a rudely made request. It should be noted that the study was implemented as a focus group discussion rather than a controlled experiment involving actual human-robot interaction. In addition, it explored a particular type of robotic technology, namely, caretaker robots explicitly programmed to support behavior modification in older adults. Therefore, it can be argued that the robot's negative response to human impoliteness was not an instance of politeness reciprocity, being instead a manifestation of a rational behavior modification strategy.

In general, the discussion in this section indicates that, while politeness and reciprocity are important research topics in HRI, there have been no previous systematic experimental studies of the interplay between robot politeness and human politeness, which is the main research focus of this paper.

---

[2] Attempts at reciprocity" were defined as "…. the child's behavior not only responding to the artifact, but expecting the artifact to respond in kind based on the child's motioning behavior, verbal directive, or offering." [17]

# 3. Method

## 3.1. Participants

Twenty-four students at a Swedish university, 9 females and 15 males, 22-38 y. o. (median age=23), took part in the study. Seventeen participants were originally from Europe, six from Asia, and one from Latin/ Central America. Participation in the study was an activity within a master's course in human-robot interaction[3]. On a scale of 1-5, the mean self-assessment score of participants' general technological skills was 4.3, experience with voice assistants was 2.9, and experience with robots was 3.2.

## 3.2. Study design

The study followed a two-factor ("Robot Politeness" X "Human Politeness") within-subject design. Four experimental conditions of the study were produced by combining two levels of the independent variable of "Robot Politeness" (*Polite* and *Impolite*, or, respectively, R-P and R-I) with two levels of the independent variable of "Human Politeness" (*Polite* and *Impolite*, or, respectively, H-P and H-I)).

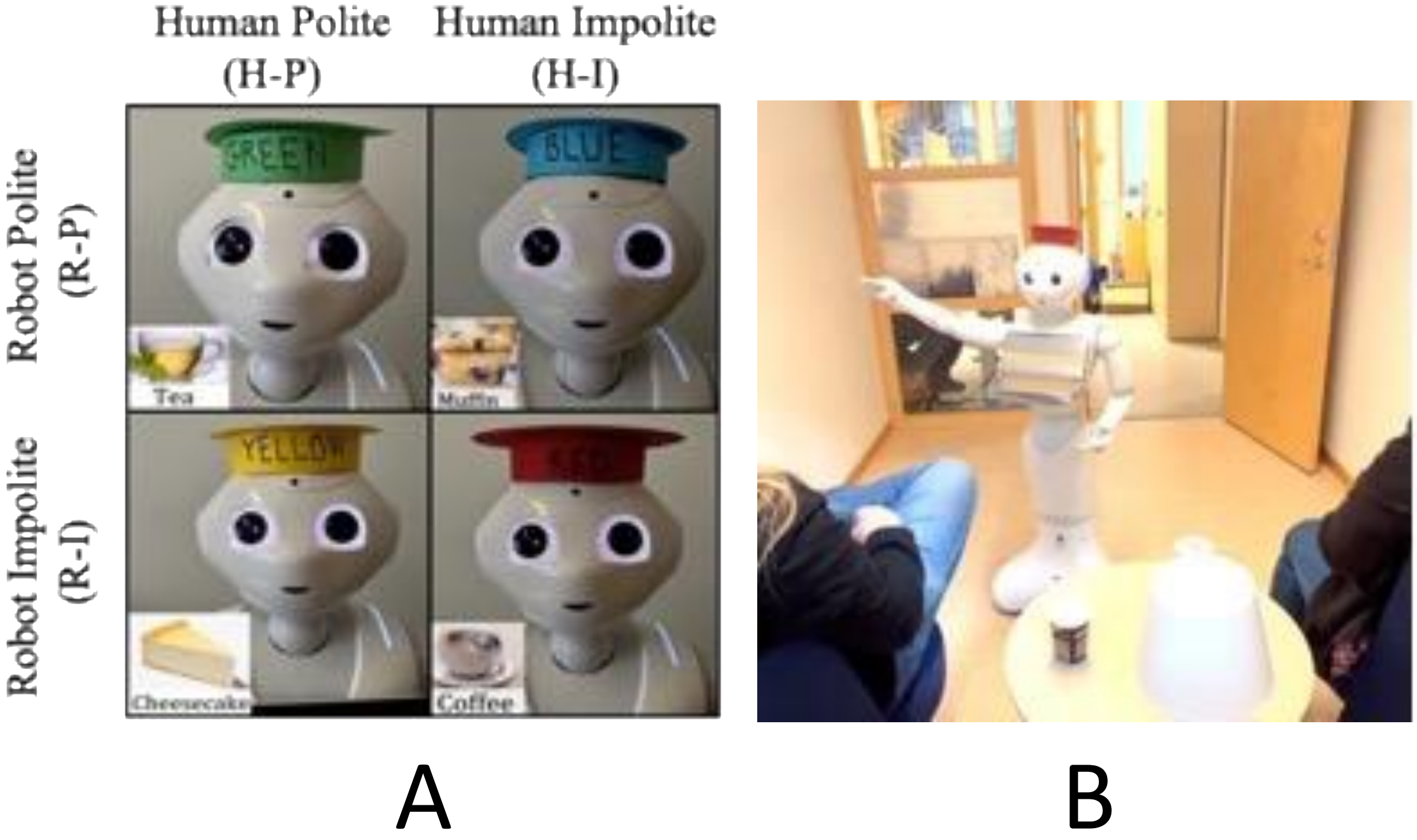

**Figure 1: A) Robot identities in the conditions of the study. B) Experimental Setting**

For each of the experimental conditions we developed a scenario involving a "waiter" robot and two customers. Four robot "identities", including two polite robots and two impolite ones were designed, one for each of the scenarios (Figure 1 a). One of the customers, an experimenter, was acting either politely or impolitely toward the robot, according to a predefined dialog script, while a participant assumed the role of a fellow customer. The roles of the polite and impolite customer were alternatively enacted by two experimenters. Each experimenter acted politely in half of the sessions and impolitely in the other half. Each participant took part in all four experimental conditions, and the order of conditions within individual sessions was balanced using the Latin Square technique.

## 3.3. Robot politeness implementation in the experimental conditions

Table 1 contains the dialogues for all scenarios[4]. The design of the scenarios was informed by existing HRI research on politeness and Brown and Levinson's notions of "positive face" and "negative face" [6]. The robots' polite verbal behavior included such phrases as "thank you" and "please" and pleasantries such as "hope you are going to enjoy your tea". The accompanying polite non-verbal behavior was mainly implemented as subtle movements such as hand rotations, bringing both hands together in front of the robot body, or bowing. The robots' impolite verbal behavior included imperative sentences such as "hurry up" and impolite utterances such as "clean up after yourselves". This language was accompanied by bold and exaggerated movements such as arms akimbo, one arm stretched out while pointing sideways and up (see Figure 1 b), and head shaking.

---

[3] The participants had the option to withdraw from the study and complete an alternative course assignment.

[4] Videos illustrating the scenarios are available at https://umeauniversity-my.sharepoint.com/:f:/g/personal/vika0001_ad_umu_se/EpAeI6IU55FMiIUQ12zlKt0BaZo1Yn8NlJag2Ns_FkObPQ

The scripts for the politely and impolitely acting humans were designed similarly. The impolite human behavior did not include any impolite postures or gestures, but the voice was raised and somewhat harsher compared to the voice for the polite scripts.

**Table 1: Dialogues in the four experimental conditions (R: robot, H: human)**

| **Impolite Robot, Impolite Human** | | **Impolite Robot, Polite Human** | |
|---|---|---|---|
| R: | *Would you like some coffee?* | R: | *Would you like some cheesecake?* |
| H: | *From you I don't expect decent coffee, but OK, bring me some of your coffee – and be fast!* | H: | *Yes, please, it would be great. Thank you very much. How long will it take?* |
| R: | *You have to wait. There are a lot of customers today, are you blind?* | R: | *It takes as long as it takes. Do you get it?* |
| The robot brings the order | | The robot brings the order | |
| R: | *Take your coffee. Now. Hurry up.* | R: | *Hurry up. Take your cheesecake. I have other customers.* |
| H: | *The coffee looks like brown water. Just as I thought. This is a pathetic place!* | H: | *Thank you very much. It looks delicious.* |
| R: | *You should leave as soon as you are finished. We have other guests waiting and they are probably much better customers than you are.* | R: | *Clean up after yourselves. Messy customers are not welcome here.* |
| **Polite Robot, Impolite Human** | | **Polite Robot, Polite Human** | |
| R: | *Would you like some muffins?* | R: | *Would you like some tea?* |
| H: | *Yeah, bring me a muffin. And if it doesn't taste good, you'll get it back. And don't make me wait forever, OK?* | H: | *Yes, please. It would be really nice. Thank you very much.* |
| R: | *Thank you for your order. I will soon be back with your muffins.* | R: | *Thank you for your order. We have excellent tea. I will bring it to you in a second.* |
| The robot brings the order | | The robot brings the order | |
| R: | *Thank you for waiting. Here are your muffins. Please take them.* | R: | *Here you are. Thank you for waiting. Please take your tea.* |
| H: | The muffins don't look good. I hope they taste better than they look. | H: | Thank you, much appreciated. It smells really nice. |
| R: | *Hope you are going to enjoy your muffins. It's great to have you with us.* | R: | *Hope you are going to enjoy your tea. It is nice to meet you and I hope you visit us frequently.* |

## 3.4. Experimental setting and procedure

The study took place in a robotics lab. After being briefed, signing a consent form, and answering several background questions, each participant took part in four enacted service scenarios. The scenarios were enacted in a room equipped with a coffee table and two chairs (Figure 1b). The participant and an experimenter, acting as customers, were sitting in the chairs while being served by a robot. The participants were asked to "play along" without actively intervening in the interaction.

In each scenario the robot first approached the customers and offered refreshments in a neutral tone. The experimenter accepted the robot's offer, either politely or impolitely. The participant also accepted the offer, mainly with "Yes, please" or "Thank you" remarks. The robot acknowledged the order, either politely or impolitely, turned away, and moved to the other side of the room, where dummy models of the refreshments were placed on the robot's chest-mounted tray. After that the robot approached the customers again, asking them to take the refreshments. The participant briefly acknowledged receiving their order, while the robot and the experimenter engaged in a brief verbal exchange, in which they were acting either politely or impolitely, according to their assigned roles. Finally, the robot turned and moved away from the customers.

After completing each scenario, the participants assessed the robot's and other person's behaviors, as well as their own feelings, by answering an online survey. When all scenarios were completed, the participants answered several concluding questions and were debriefed about the study. The debriefing included informing the participants about the "robot-experimenter" dialogs being scripted and using the Wizard of Oz approach to control the robots. The briefing, data collection, and debriefing took place in a separate office and were conducted and supervised by a researcher. The entire session was approximately 40-45 minutes long.

## 3.5. Equipment and software

A Pepper robot was employed in the study. The robot's behaviors were remotely controlled with a Wizard of Oz interface WoZ4U [31]. The Wizard sat in an adjacent room but could hear what was said during the interaction, and see somewhat through a frosty window (see Figure 1b). This made it easier for the Wizard to control the forward movements and rotations of the robot as well as the timing of the robot's verbal utterances. All of the robot's utterances and accompanying gestures were coded in YAML files[5]. The pitch and speed of speech were kept similar across all four scenarios.

## 3.6. Data collection

Data collection was performed using an online survey comprising the following sections: "Background", "Post-condition" (identical sections for each of the four conditions of the study), and "Concluding questions". The "Background" section contained questions about the participants' age, gender, and technological skills. Each of the "Post-condition" sections included three Likert-type scales, each comprising several 5-point Likert items (from "strongly disagree" to "strongly agree", with an additional "not applicable" option) for assessing the following attributes:

- "Robots' behavior" (*Socially acceptable/ Polite/ Friendly/ Justifiable/ Fair/ Responsive/ Competent/ Humanlike*),
- "The other person's behavior" (*Socially acceptable/ Polite/ Friendly/ Justifiable/ Fair*), and
- "Situation", that is, participant's subjective experiences of the situation (*Comfortable/ Safe/ Relaxed*).

The development of the scales was influenced by existing HRI instruments, Goodspeed [4] and RoSAS [7]. However, since these instruments do not specifically deal with politeness, we chose to construct our own set of scales, which allowed us to address several politeness-related aspects. It should be noted that our aim was not to create a new instrument, and the application scope of the scales was limited to collecting data in this particular study.

The "Concluding questions" section was comprised of 5-point Likert items and free-text responses dealing with the participant's overall experience of the study conditions. For each of the four conditions the participants were asked to describe whether they: (a) liked the robots, (b) empathized with the robots (i.e., understood their "feelings"), and (c) empathized with the other person (an experimenter). The questions were intentionally formulated rather generally to elicit the participants' own interpretations of "liking" and "empathizing".

In addition, the participants were asked to provide their general opinions about a human being rude to a rude robot, and a robot being rude to a rude human. In each case, four Likert items (*Acceptable/ Understandable/ Excusable/ Normal*) and a free-text description were used.

## 3.7. Ethical aspects

The project was submitted for review to the Swedish Ethical Review Authority (application 2022-04747-01) and was found to not require ethical review under Swedish legislation (2003:615). The study was conducted according to national and institutional ethical research standards, including informed consent and data confidentiality protection. A potential negative impact of impolite robots' behavior on the participants was mitigated by the participants being observers of impolite exchanges between a robot and a researcher, rather than persons, actively engaged in the exchanges. During the study, the participants were always treated politely by other people. The participants were provided with the option of withdrawing from the study and completing an alternative course assignment. At post-session debriefings the participants were informed that the dialogs enacted in the study were scripted and that a Wizard of Oz approach was used to control the robots.

# 4. Results

## 4.1. Post-Condition Questionnaires

### 4.1.1. Statistical Analysis and Background

The post-condition questions are analyzed using two-way repeated measure ANOVA, where the two factors are human-polite/impolite and robot-polite/impolite. Since there are only two levels on each factor, a test of sphericity is not applicable. The effect of the experimenter acting as the other human is tested with one-way ANOVA.

There were no statistically significant differences between the two alternative assignments of the roles of the polite and the impolite humans to the experimenters (who switched roles after half the sessions). There were no statistically significant differences between participants from Europe, Asia, and South/ Central America. In general, technological skills, experience with voice assistants, and experience with robots did not significantly affect the participant's responses. However, a small positive correlation exists between overall technological skills and the robot's behavior perceived as justifiable ($r=0.220$, $p=0.031$) and humanlike ($r=0.225$, $p=0.35$). Finally, experience with voice assistance correlated with robot responsiveness ($r=0.252$ $p=0.020$). While these correlations are small, it is worth noting that the population used in this study was highly homogeneous in their general technology experience and had prior experience in programming the Pepper robot.

[5] The YAML files will be made available on request.

All between-condition questions except for the robots' human likeness yielded statistically significant effects between the four conditions. For every between-condition measurement, a two-way repeated measure ANOVA was performed on the two independent variables, robot politeness and human politeness. Sphericity tests are not applicable because there are only two levels per factor.

### 4.1.2. Experience of the robots' and the other person's behavior

Table 2 shows the means and standard deviation of both the Robots' and the Other Person for both factors. The politeness experienced by the robot and the human was consistent with the conditions of the study. The robot in the polite conditions had an average score in politeness of 4.88 and impolite 1.56 (3.32 in difference). Similarly, the human scored 4.73 when polite and 2.08 when impolite. Both effects are statistically significant (two-way repeated measures ANOVA, Robot F=877.018, p<0.001, Human F= 239.937, p<0.001, no significant interaction effect, see Table 3 & Table 4).

**Table 2: Descriptive Statistics of Robots' and Other Persons behavior**

| | Robot | Human | | Polite | Socially acceptable | Friendly | Justifiable | Fair | Responsive | Competent | Humanlike |
|---|---|---|---|---|---|---|---|---|---|---|---|
| **Robots' Behavior** | Impolite | Impolite | Mean | 1.71 | 2.50 | 1.71 | 3.21 | 3.12 | 3.57 | 2.82 | 3.18 |
| | | | σ | .751 | 1.180 | .859 | 1.215 | 1.191 | 1.121 | 1.140 | 1.259 |
| | | Polite | Mean | 1.42 | 2.00 | 1.58 | 1.87 | 1.83 | 2.33 | 2.32 | 2.68 |
| | | | σ | .717 | 1.022 | .974 | .850 | 1.007 | 1.065 | 1.086 | .995 |
| | | **Total** | Mean | 1.56 | 2.25 | 1.65 | 2.54 | 2.48 | 2.95 | 2.57 | 2.93 |
| | | | σ | .741 | 1.120 | .911 | 1.237 | 1.271 | 1.248 | 1.129 | 1.149 |
| | Polite | Impolite | Mean | 4.79 | 4.50 | 4.54 | 4.12 | 3.83 | 2.95 | 4.05 | 2.82 |
| | | | σ | .588 | .780 | .779 | 1.116 | 1.239 | 1.214 | .899 | 1.181 |
| | | Polite | Mean | 4.96 | 4.58 | 4.71 | 4.63 | 4.63 | 4.14 | 4.50 | 3.55 |
| | | | σ | .204 | .654 | .624 | .495 | .875 | .854 | .740 | 1.057 |
| | | **Total** | Mean | 4.88 | 4.54 | 4.63 | 4.37 | 4.23 | 3.53 | 4.27 | 3.18 |
| | | | σ | .444 | .713 | .703 | .890 | 1.134 | 1.202 | .845 | 1.167 |
| | **Total** | Impolite | Mean | 3.25 | 3.50 | 3.12 | 3.67 | 3.48 | 3.26 | 3.43 | 3.00 |
| | | | σ | 1.695 | 1.414 | 1.645 | 1.243 | 1.255 | 1.197 | 1.189 | 1.220 |
| | | Polite | Mean | 3.19 | 3.29 | 3.15 | 3.25 | 3.23 | 3.24 | 3.41 | 3.11 |
| | | | σ | 1.864 | 1.557 | 1.774 | 1.551 | 1.692 | 1.322 | 1.436 | 1.104 |
| | | **Total** | Mean | 3.22 | 3.40 | 3.14 | 3.46 | 3.35 | 3.25 | 3.42 | 3.06 |
| | | | σ | 1.772 | 1.483 | 1.702 | 1.414 | 1.487 | 1.253 | 1.311 | 1.158 |
| **Other Persons' Behavior** | Impolite | Impolite | Mean | 1.88 | 1.42 | 1.38 | 1.79 | 1.96 | | | |
| | | | σ | .741 | .504 | .495 | .721 | .878 | | | |
| | | Polite | Mean | 4.58 | 4.75 | 4.67 | 4.17 | 4.21 | | | |
| | | | σ | .929 | .847 | .917 | .984 | 1.215 | | | |
| | | **Total** | Mean | 3.23 | 3.08 | 3.02 | 2.96 | 3.11 | | | |
| | | | σ | 1.601 | 1.820 | 1.816 | 1.474 | 1.550 | | | |
| | Polite | Impolite | Mean | 2.29 | 1.87 | 1.83 | 2.00 | 2.04 | | | |
| | | | σ | 1.197 | 1.035 | 1.049 | 1.022 | .999 | | | |
| | | Polite | Mean | 4.87 | 5.00 | 4.87 | 4.79 | 4.67 | | | |
| | | | σ | .338 | .000 | .338 | .509 | .702 | | | |
| | | **Total** | Mean | 3.58 | 3.44 | 3.35 | 3.40 | 3.35 | | | |
| | | | σ | 1.569 | 1.737 | 1.720 | 1.621 | 1.578 | | | |
| | **Total** | Impolite | Mean | 2.08 | 1.65 | 1.60 | 1.90 | 2.00 | | | |
| | | | σ | 1.007 | .838 | .844 | .881 | .933 | | | |
| | | Polite | Mean | 4.73 | 4.87 | 4.77 | 4.49 | 4.44 | | | |
| | | | σ | .707 | .606 | .692 | .831 | 1.009 | | | |
| | | **Total** | Mean | 3.41 | 3.26 | 3.19 | 3.18 | 3.23 | | | |
| | | | σ | 1.587 | 1.778 | 1.767 | 1.557 | 1.561 | | | |

Human- and Robot Politeness affected all but one post-condition parameter: the robot's human likeness. Thus, the robot and the human were seen as significantly more socially acceptable, friendly, justifiable, and fair in polite conditions (see Table 2 and Table 4 for significance). In addition, the robot was seen as increasingly competent and responsive in the polite condition.

**Table 3: A Two-way repeated ANOVA for the robots' behavior**

| Source | Measure | Type III Sum of Squares | df | Mean Square | F | Sig. | Partial Eta Squared |
|---|---|---|---|---|---|---|---|
| **Robot** | SociallyAccepted | 116.679 | 1 | 116.679 | 132.805 | <.001 | .869 |
| | Friendly | 186.012 | 1 | 186.012 | 316.937 | <.001 | .941 |
| | Competent | 65.190 | 1 | 65.190 | 49.557 | <.001 | .712 |
| | Justifiable | 72.429 | 1 | 72.429 | 75.955 | <.001 | .792 |
| | Fair | 61.714 | 1 | 61.714 | 51.892 | <.001 | .722 |
| | Responsive | 8.048 | 1 | 8.048 | 9.783 | .005 | .328 |
| | HumanLike | 1.440 | 1 | 1.440 | .689 | .416 | .033 |
| **Error (Robot)** | SociallyAccepted | 17.571 | 20 | .879 | | | |
| | Friendly | 11.738 | 20 | .587 | | | |
| | Competent | 26.310 | 20 | 1.315 | | | |
| | Justifiable | 19.071 | 20 | .954 | | | |
| | Fair | 23.786 | 20 | 1.189 | | | |
| | Responsive | 16.452 | 20 | .823 | | | |
| | HumanLike | 41.810 | 20 | 2.090 | | | |
| **Human** | SociallyAccepted | 2.679 | 1 | 2.679 | 2.483 | .131 | .110 |
| | Friendly | .012 | 1 | .012 | .016 | .900 | .001 |
| | Competent | .048 | 1 | .048 | .148 | .705 | .007 |
| | Justifiable | 5.762 | 1 | 5.762 | 7.322 | .014 | .268 |
| | Fair | 3.048 | 1 | 3.048 | 3.133 | .092 | .135 |
| | Responsive | .048 | 1 | .048 | .058 | .812 | .003 |
| | HumanLike | .298 | 1 | .298 | .543 | .470 | .026 |
| **Error (Human)** | SociallyAccepted | 21.571 | 20 | 1.079 | | | |
| | Friendly | 14.738 | 20 | .737 | | | |
| | Competent | 6.452 | 20 | .323 | | | |
| | Justifiable | 15.738 | 20 | .787 | | | |
| | Fair | 19.452 | 20 | .973 | | | |
| | Responsive | 16.452 | 20 | .823 | | | |
| | HumanLike | 10.952 | 20 | .548 | | | |
| **Robot * Human** | SociallyAccepted | 3.440 | 1 | 3.440 | 7.015 | .015 | .260 |
| | Friendly | .964 | 1 | .964 | 1.636 | .215 | .076 |
| | Competent | 5.762 | 1 | 5.762 | 9.047 | .007 | .311 |
| | Justifiable | 17.190 | 1 | 17.190 | 33.349 | <.001 | .625 |
| | Fair | 21.000 | 1 | 21.000 | 18.667 | <.001 | .483 |
| | Responsive | 29.762 | 1 | 29.762 | 23.127 | <.001 | .536 |
| | HumanLike | 7.440 | 1 | 7.440 | 11.617 | .003 | .367 |
| **Error( Robot * Human)** | SociallyAccepted | 9.810 | 20 | .490 | | | |
| | Friendly | 11.786 | 20 | .589 | | | |
| | Competent | 12.738 | 20 | .637 | | | |
| | Justifiable | 10.310 | 20 | .515 | | | |
| | Fair | 22.500 | 20 | 1.125 | | | |
| | Responsive | 25.738 | 20 | 1.287 | | | |
| | HumanLike | 12.810 | 20 | .640 | | | |

**Table 4: Repeated measure two-way ANOVA for the other person's behavior**

| Source | Measure | Type III Sum of Squares | df | Mean Square | F | Sig. | Partial Eta Squared |
|---|---|---|---|---|---|---|---|
| Robot | Socially Accepted | 2.909 | 1 | 2.909 | 4.495 | .046 | .176 |
| | Friendly | 2.557 | 1 | 2.557 | 4.070 | .057 | .162 |
| | Justifiable | 3.682 | 1 | 3.682 | 7.493 | .012 | .263 |
| | Fair | 1.636 | 1 | 1.636 | 3.316 | .083 | .136 |
| Error (Robot) | Socially Accepted | 13.591 | 21 | .647 | | | |
| | Friendly | 13.193 | 21 | .628 | | | |
| | Justifiable | 10.318 | 21 | .491 | | | |
| | Fair | 10.364 | 21 | .494 | | | |
| Human | Socially Accepted | 158.227 | 1 | 158.227 | 99.865 | <.001 | .826 |
| | Friendly | 225.920 | 1 | 225.920 | 170.478 | <.001 | .890 |
| | Justifiable | 152.909 | 1 | 152.909 | 127.978 | <.001 | .859 |
| | Fair | 132.545 | 1 | 132.545 | 78.508 | <.001 | .789 |
| Error (Human) | Socially Accepted | 33.273 | 21 | 1.584 | | | |
| | Friendly | 27.830 | 21 | 1.325 | | | |
| | Justifiable | 25.091 | 21 | 1.195 | | | |
| | Fair | 35.455 | 21 | 1.688 | | | |
| Robot * Human | Socially Accepted | .182 | 1 | .182 | .604 | .446 | .028 |
| | Friendly | .557 | 1 | .557 | 3.662 | .069 | .148 |
| | Justifiable | .727 | 1 | .727 | 2.100 | .162 | .091 |
| | Fair | .727 | 1 | .727 | .939 | .344 | .043 |
| Error (Robot * Human) | Socially Accepted | 6.318 | 21 | .301 | | | |
| | Friendly | 3.193 | 21 | .152 | | | |
| | Justifiable | 7.273 | 21 | .346 | | | |
| | Fair | 16.273 | 21 | .775 | | | |

### 4.1.3. Experience of the Situation

The participant's experience of the situation was affected positively by the politeness of the robot and the other human. When both the robot and the other person were polite, the participant felt most comfortable, safe, and relaxed. The opposite was true when both were impolite. The difference between the independent variables is significant (see Table 5), and the pairwise comparison between all dependent variables is significant, $p<0.005$). Thus, situational parameters depend on both actors, while the robot's influence on the situation is slightly bigger than the human's: relaxed and comfortable.

**Table 5: Descriptive statistics of situation-parameters**

| Robot | Human | | Comfortable | Safe | Relaxed |
|---|---|---|---|---|---|
| Impolite | Impolite | Mean | 2.3750 | 2.8750 | 2.4583 |
| | | N | 24 | 24 | 24 |
| | | $\sigma$ | 1.01350 | 1.22696 | 1.21509 |
| | Polite | Mean | 2.8750 | 3.5833 | 3.0000 |
| | | N | 24 | 24 | 24 |
| | | $\sigma$ | 1.26190 | 1.28255 | 1.28537 |
| | Total | Mean | 2.6250 | 3.2292 | 2.7292 |
| | | N | 48 | 48 | 48 |
| | | σ | 1.16006 | 1.29220 | 1.26726 |
| Polite | Impolite | Mean | 3.1667 | 3.7917 | 3.2083 |
| | | N | 24 | 24 | 24 |
| | | σ | 1.20386 | 1.17877 | 1.14129 |
| | Polite | Mean | 4.2083 | 4.6667 | 4.1667 |
| | | N | 24 | 24 | 24 |
| | | σ | .72106 | .63702 | .76139 |
| | Total | Mean | 3.6875 | 4.2292 | 3.6875 |
| | | N | 48 | 48 | 48 |
| | | σ | 1.11386 | 1.03635 | 1.07498 |
| Total | Polite | Mean | 2.7708 | 3.3333 | 2.8333 |
| | | N | 48 | 48 | 48 |
| | | σ | 1.17128 | 1.27719 | 1.22619 |
| | Impolite | Mean | 3.5417 | 4.1250 | 3.5833 |
| | | N | 48 | 48 | 48 |
| | | σ | 1.21967 | 1.14157 | 1.19988 |
| | Total | Mean | 3.1562 | 3.7292 | 3.2083 |
| | | N | 96 | 96 | 96 |
| | | σ | 1.25092 | 1.26889 | 1.26422 |

In Table 6, one can see that all the parameters are affected by the conditions individually and that there is an interaction between the two independent variables and how comfortable the participant is (Human*Robot F=4.267 p=0.05). Thus, “Comfortable” is affected by the combination of human and robot politeness, although lower than the independent variables in isolation. The only significant order effect was that participants' comfort increased with the rounds (Person r = 0.236, p = 0.21).

**Table 6: Two-Way Repeated Measure ANOVA for situation-parameters**

| Source | Measure | Type III Sum of Squares | df | Mean Square | F | Sig. | Partial Eta Squared |
|---|---|---|---|---|---|---|---|
| Human | Relaxed | 22.042 | 1 | 22.042 | 22.573 | <.001 | .495 |
| | Safe | 24.000 | 1 | 24.000 | 23.489 | <.001 | .505 |
| | Comfortable | 27.094 | 1 | 27.094 | 23.824 | <.001 | .509 |
| Error (Human) | Relaxed | 22.458 | 23 | .976 | | | |
| | Safe | 23.500 | 23 | 1.022 | | | |
| | Comfortable | 26.156 | 23 | 1.137 | | | |
| Robot | Relaxed | 13.500 | 1 | 13.500 | 28.227 | <.001 | .551 |
| | Safe | 15.042 | 1 | 15.042 | 27.769 | <.001 | .547 |
| | Comfortable | 14.260 | 1 | 14.260 | 25.250 | <.001 | .523 |
| Error (Robot) | Relaxed | 11.000 | 23 | .478 | | | |
| | Safe | 12.458 | 23 | .542 | | | |
| | Comfortable | 12.990 | 23 | .565 | | | |
| Human * Robot | Relaxed | 1.042 | 1 | 1.042 | 2.091 | .162 | .083 |
| | Safe | .167 | 1 | .167 | .338 | .567 | .014 |
| | Comfortable | 1.760 | 1 | 1.760 | 4.267 | .050 | .156 |
| Error (Human * Robot) | Relaxed | 11.458 | 23 | .498 | | | |
| | Safe | 11.333 | 23 | .493 | | | |
| | Comfortable | 9.490 | 23 | .413 | | | |

#### 4.1.4. Interaction Effects

Two-way repeated measure ANOVA over the post-condition Likert scale data of the robot and other person shows a significant interaction ($p<0.05$) effect between "Human politeness" and "Robot Politeness" in six of the dependent variables measuring the robot's behavior, Socially Accepted ($F=7.015$, $p=0.15$), Competent ($F=9.047$, $p=0.001$), Justifiable ($F=33.349$, $p<0.001$), Fair ($F=18.667$, $p<0.001$), Responsive ($F=23.127$, $p<0.001$), and Human Like ($F=11.617$, $p=0.003$). Thus, this means that the other person's politeness affects how the robot is experienced. For the variables responsiveness and human likeness, the interaction effect is higher than that of robot politeness in isolation, especially for human likeness, where Robot Politeness shows no significant effect. The robot's behavior was assessed as most justifiable when both the human and the robot were polite and least justifiable when the human was polite and the robot was impolite. Similar patterns were found for fairness, responsiveness, and competence.

In the R-I/H-P condition, the other human's behavior was seen as less fair and less justifiable than the robot's behavior (see Table 2). Furthermore, there is an indication that when the human is polite, and the robot is impolite, the human's actions appear less justifiable. However, the differences are small, and it is hard to draw firm conclusions due to the difference in roles.

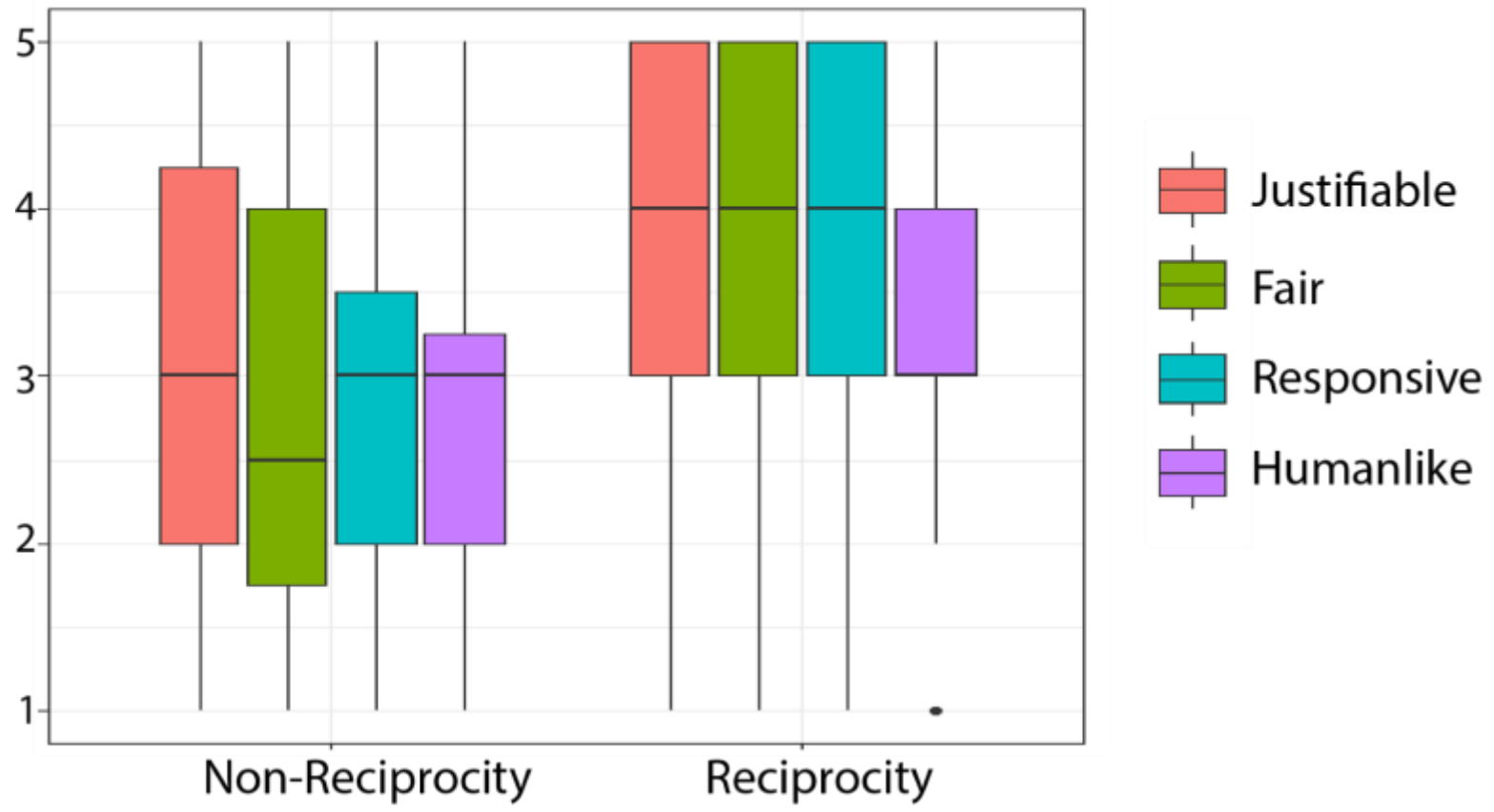

**Figure 2: Robots assessment in reciprocity vs. non-reciprocity conditions.**

In the scenarios where the robot and human reciprocated their behavior, i.e. R-I/H-I and R-P/H-P, the robot had higher scores on the scales of *justifiable*, *fair*, and *responsive* (see Figure 2). Reciprocity did not significantly affect how the human or the situation was perceived. Notably, there are indications that the robot was regarded as more humanlike in the reciprocity situations ($p=.012$, 0,6136 in mean difference, $z=2.521$, $p=0.0117$).

#### 4.1.5. Correlation between post-condition questions

All the parameters within a category (i.e., "Robots' behavior", "The other person's behavior", and "Situation") significantly positively correlate with each other ($p<0.05$), except for robot politeness, which does not correlate with human likeness. All the parameters of robots' behavior, except for human likeness and other persons' behavior, positively correlate with all the situation parameters ($p<0.05$). Human likeness only correlates with the situation parameter "comfortable."

### 4.2. Concluding questions: Comparative assessments of the study conditions

The questions of how/whether the participants, in each of the four experimental conditions, *liked the robot*, *felt empathy for the robot*, and *felt empathy for the other human* were answered by providing Likert scale scores and free-text notes.

#### 4.2.1. Robots' likeability

Likert scale scores (Figure 3) showed a clear difference between polite and impolite robots. Both polite robots, *Green* (R-P/H-P) and *Blue* (R-P/H-I), were generally assessed as positive, while both impolite ones, *Red* (R-I/H-I) and *Yellow* (R-I/H-P), were generally assessed as negative. Mean scores also indicate a certain preference of reciprocity over non-reciprocity conditions for both polite robots (*Green*, 4.4, over *Blue*, 3.8) and impolite robots (*Red*, 2.3, over *Yellow*, 2.1).

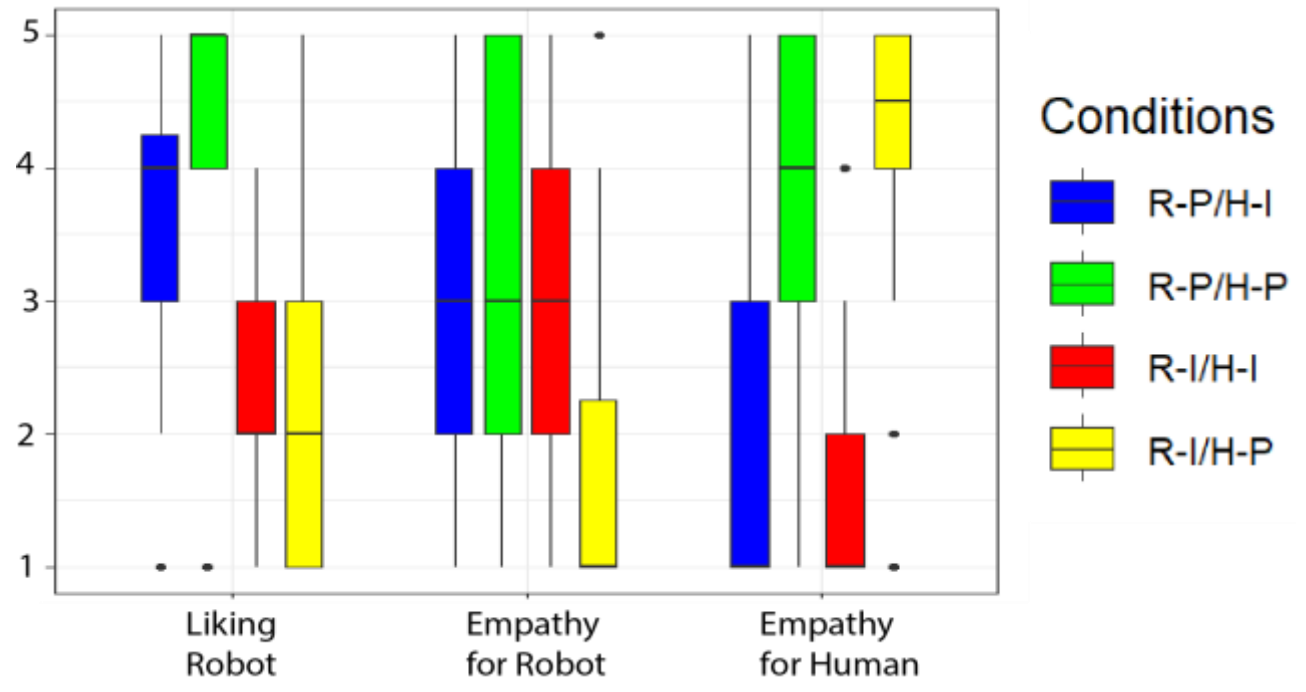

**Figure 3: Participants' assessments of "liking the robot" "feeling empathy for the robot", and "feeling empathy for the other human" (R: robot, H: human; P: polite, I: impolite).**

The most common ranking of the robots in the free-text responses, from most liked to most disliked, was consistent with the Likert scale data. In 11 out of 18 complete responses the robots were generally ranked in the following order: *Green* (R-P/H-P), *Blue* (R-P/H-I), *Red* (R-I/H-I), and *Yellow* (R-I/H-P). For instance, participant P16 observed: "*The red robot was rude, but so was the other customer, so I kind of understand it. The green robot was perfectly adequate. The yellow robot was rude but the other customer was polite so I disliked this one the most. For the blue robot, nothing stood out too much.*" (P16).

Table 7 shows a more detailed overview of participants' free-text responses by presenting all instances of explicit robot assessment as coded according to the following 7-point ordinal scale: *The best/ Nice/ (Nice but) something is off/Not bad/ (Not nice but) understandable/ Not nice/ The worst*.

**Table 7: Coded free-text participants' assessments of the robots (R: robot, H: human; P: polite, I: impolite)**

| | The best | Nice | "some-thing is off" | Not bad | under-standable | Not nice | The worst |
|---|---|---|---|---|---|---|---|
| R-P/H-P (Green ) | 7 | 6 | 3 | 0 | 0 | 0 | 0 |
| R-P/H-I (Blue) | 1 | 8 | 4 | 3 | 0 | 0 | 0 |
| R-I/H-I (Red) | 0 | 2 | 1 | 0 | 10 | 5 | 1 |
| R-I/H-P (Yellow) | 0 | 1 | 1 | 2 | 0 | 11 | 3 |

The overall pattern of the comparative assessment of the robots, shown in Table 7, is consistent with the Likert scale data, and the most frequent ranking order described above. The table also highlights several additional aspects. First, a common perception of the *Red* (R-I/H-I) robot was "not nice but understandable", which differentiates it from all other robots. Second, some participants perceived polite robots not entirely positively ("nice but something was off"), mostly because they thought the robots were "too polite". Finally, there was a wide range of individual assessments of impolite robots. For instance, the *Red* (R-I/H-I) robot was disliked the most by one participant ("a *combination of unpolite person and a rude robot was the worst*", P10), but two other participants described the same robot as the second best (e.g., "*I also liked the red robot standing up for itself*", P23).

### 4.2.2. Empathy for robots

According to the Likert scale data, *Yellow* (R-I/ H-P), with a score of 1.9, was the only robot receiving a below-average mean score on "feeling empathy for". The polite *Blue* robot (R-P/H-I) was second worst, with a nearly average score of 2.9. Both *Green* (R-P/H-P) and *Red* (R-I/H-I) received positive scores: respectively, 3.4 and 3.1.

Free-text responses of most participants (all but 5) showed some form of empathy for the robots. Several factors affecting empathy were reported, sometimes in the same response. *Reciprocity* (referred to as, e.g., "fairness") was mentioned in 12 responses, either by its presence (e.g., "*The robots that were fair, I felt more empathy for",* P07) or absence (e.g., "*If the other customer was nice, I had less empathy for the robot if it was mean"*, P01). A special case of a lack of reciprocity was a somewhat negative attitude toward being too nice to rude people: "*It feels weird if a robot ignores rude people and is too friendly. If the customer is also friendly then it is normal.*" (P14).

Another reason for empathy, according to 7 responses, was compassion, feeling sorry for the robot: "*I felt most empathetic towards the blue robot as it was nice but the human customer was rude regardless which made me feel bad for it.*" (P03).

"And the red one seemed like they just had a bad day and so I empathize with that, especially having worked in customer service myself in the past." (P23).

Yet another reason was *politeness*: 6 participants pointed out that they felt more empathy for polite robots, e.g.: "*I felt more empathy towards the robots that were nice." (P08).*

Five participants noted that their attitude toward robots could not be described in terms of empathy. Three of them observed that they did not have any feelings for the robots (e.g., "*For all, since I don't think the robot has feelings I don't care if the customer was rude or not*", P11). Two participants mentioned that they understood the robots' *behavior*, rather than its *feelings*, e.g.:

"*I didn't really 'feel' much for the robot, but I kind of understood the red robot as it answered impolite while the other customer was impolite as well*." (P16).

#### 4.2.3. Empathy for the other person

The Likert scale data (Figure 3) shows a clear difference between feeling empathy for a polite human (mean scores 3.7 and 4.1) vs. an impolite human (mean scores 1.9 and 1.7). The maximum empathy was experienced in the scenario in which a polite human was treated impolitely (R-I/ H-P, featuring the *Yellow* robot).

In their free-text responses, the majority of participants reported empathy for the other human, and the responses were consistent with the Likert scale data: the participants generally felt more empathy for persons who acted politely, especially when they were treated impolitely by the robot.

"*I felt more empathy towards the "nice" person, and felt it the most when the robot was being somewhat rude.*" (P08)

"*As for the yellow robot, I think I felt sorry for the other customer as they were very nice and the robot wasn't.*" (P14).

An exception was a response reporting a lack of empathy in the "polite human/ impolite robot" condition:

"*I didn't feel much for the person in the scenario with the yellow robot because the situation felt kind of unnatural and tense*." (P04)

Finally, three participants noted they had no empathy for the other human, because they either did not pay enough attention to the person (2 responses) or did not take the situation seriously (1 response).

### 4.3. Concluding questions: General reflections

#### 4.3.1. Human being rude to a rude robot

The participants used Likert scale items and free-text responses to express their opinions about a person being rude to a rude robot. The Likert scale data (Figure 4) reflects a wide range of assessments, with median scores being near-average for the scales of "Acceptable", Understandable, and "Excusable", and below-average for "Normal".

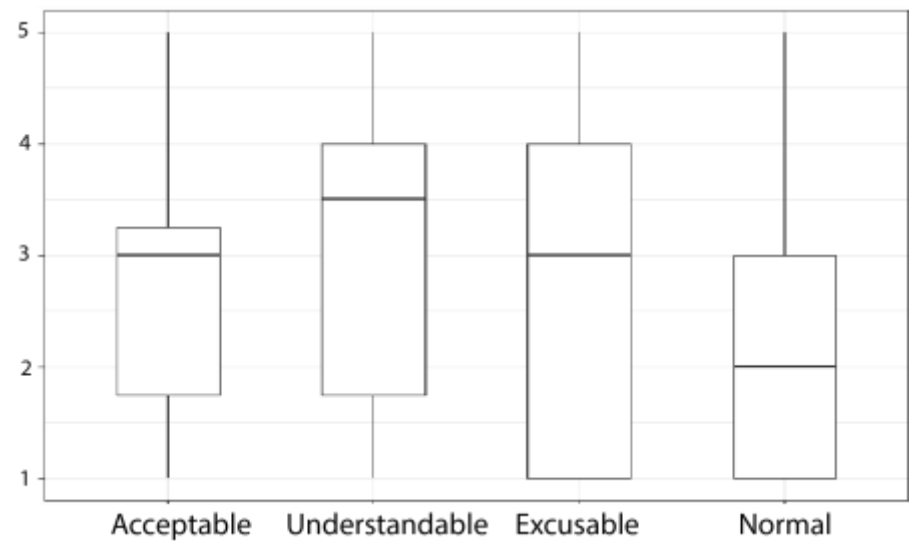

**Figure 4. Assessment of a human being rude to a rude robot.**

In their free-text responses, the majority of the participants (20 out of 24) expressed a generally negative attitude toward people being rude to rude robots. The main reason was that such behavior may negatively affect other people and the person himself/herself, either directly or potentially, e.g.: "*I do not find it socially acceptable to be rude to a robot despite it not having emotions, because there are still people around to hear the interaction and to get the sense that the person is not a nice person in general*." (P23).

"*It's probably not good for the own personality development of the human. <...> It will probably become a common behavior for you if you do it often enough*." (P02).

Twelve participants also mentioned that being rude to rude robots may be, to a certain extent, understandable, and in some cases even acceptable. Some of the reasons given for this were because robots have no feelings, are rude themselves, are not good at certain tasks, can learn from feedback and thus can avoid being rude in the future, and because it is better to be rude toward a robot than a human.

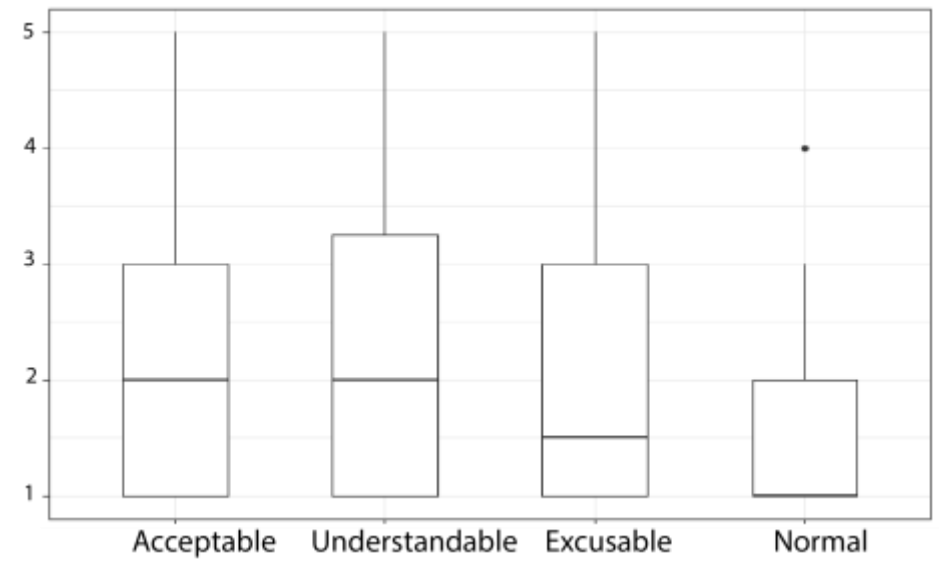

**Figure 5: Assessment of a robot being rude to a rude human.**

#### 4.3.2. Robot being rude to a rude human

The Likert scale data shows that rude robots' behavior toward humans was assessed more negatively than rude humans' behavior toward robots (Figure 5). Median scores were the lowest for the scales of "Normal" and "Excusable", and somewhat higher, but still negative, for the scales of "Acceptable" and "Understandable".

In free-text responses, 17 participants were either completely or partly negative about robots being rude to rude humans. The main reason, cited in the responses, was, essentially, that robots do not have feelings and therefore have no right to act as if they are offended. "*I think robots should not be excused for rude behaviour towards humans as they themselves do not have the same emotional capabilities as humans do, and as such the rude encounter will not affect them at all while it may cause lasting effects on humans*" (P03). "*robots don't have feelings and should therefore not be rude even if a human is*" (P11).

In four generally negative responses it was pointed out that in some cases rude robots' behavior may be justified, for instance, robots may serve as "guards or something similar" or rudeness can be a way to send people necessary social signals.

Finally, six participants thought that rude robot's behavior can be acceptable or understandable, because

- a robot may act improperly due to its limited knowledge about the social context (e.g., "*I would say I have more openness towards a robot because it does not know all social cues and codes. This makes it more okay for a robot servant to be mean to a customer than the other way around*", P01),
- rude behavior should in general be discouraged (e.g., "*I do feel very much the same way about the robot as I did about the human in the question above, and I do feel like rude behaviour could be expected if the robot is treated badly by the human*", P04),
- robots should not be treated as "slaves" ("*I think the robot should also get the same treat as human. We shouldn't make robot as a "slave". If you want others (no matter human or robot) be polite to you, you should first be pilot to others"*, P13), and,
- as in case of humans, one can expect robots to be rude when they are treated badly ("*I think that for a robot it is more acceptable to be rude to a rude customer than for a human server to be rude, otherwise people might think it is okay to speak like that because the robot cannot do anything*", P14).

## 5. Discussion

The study reported in this paper examined the effect of the human treatment of a robot (polite or impolite) on the perception of the robot's behaviour. The following hypotheses were tested: H1. *Social attributes of a robot are perceived differently depending on whether the robot is treated politely or impolitely,* and H2: *Human assessments of a robot's behavior are affected by an expectation of politeness reciprocity*. This section discusses the empirical evidence collected in the study in order to assess the hypotheses, as well as the methodology we employed.

### 5.1. Politeness implementation in the study conditions

Human-robot interactions in the experimental conditions of the study were designed to give an observer an impression of a polite or an impolite behaviour exhibited by the human and the robots. Two potential issues were associated with this study design. First, there was a risk that we did not manipulate the experimental conditions successfully enough. The perception of politeness is a subjective experience, which, as any such experience, cannot be controlled directly. Therefore, there was a possibility that the participants would not see the intended difference between a polite and an impolite behaviour of the robots, the humans, or both. The empirical evidence obtained in the study indicates that this risk has been avoided. Statistically significant main effects of each of the independent variables, "Robot politeness" and "Human politeness", on the post-condition scores of "*politeness*" (as well as on all other scores, except for robots' "*human-likeness*") show that politeness was successfully manipulated in the experimental conditions of the study.

Second, the participants in the study were mostly observing another person interacting with a robot, rather than themselves being active participants in the interaction. There was a potential risk that the participants would adopt the perspective of indifferent bystanders, not paying much attention to the human-robot interactions around them. The evidence from the study shows that this risk was avoided, too. Participants' free-text responses indicate that, despite their "included observer" position, most of the participants were cognitively and emotionally engaged in the human-robot interactions taking place in the conditions of the study. Only three participants mentioned having no feelings for the robots (two additional participants noted that they understood the robots' "behavior" rather than "feelings") and only four participants did not have any empathy for the human.

## 5.2. Effect of human politeness on social perception of robot's behavior

As mentioned in the "Results" section, we found a significant interaction effect of the independent variables, "Robot politeness" and "Human politeness" on the participants' post-condition scores, provided for the following "Robots' behaviour" Likert scale items: *"Justifiable", "Competent", "Responsive", and "Fair"*. This finding shows that participants' perception of the respective social attributes of a robot's behavior was affected by how the robot was treated by a human: the perception depended on whether the robots were treated politely or impolitely, and, in particular, an impolite robot was perceived as less negative if the human was impolite. In other words, the effect of human (im)politeness on the participants' assessments of a robot's behaviour was not straightforward: it cannot be described as only positive or only negative. Instead, the robots' behavior was assessed more positively when it "mirrored" a polite or an impolite human behavior. Therefore, these post-condition data provide a partial confirmation of the first hypothesis, H1: "*Social attributes of a robot are perceived differently depending on whether the robot is treated politely or impolitely*". The data indicates that the hypothesis is valid for at least *some* social attributes.

Additional evidence supporting the hypothesis comes from the overall pattern of data in the comparative assessments of study conditions in "Concluding questions". The pattern was consistent with the post-condition data: while both impolite robots employed in the study were generally assessed as negative in terms of likeability and empathy, the robot that was treated impolitely was assessed more positively compared to the one that was impolite to a polite human.

Several aspects of the quantitative analysis deserve special attention. First, we did not find a statistically significant interaction effect of the independent variables on the scores of "*Socially acceptable*", "*Friendly*", "*Polite*", and "*Human-like*". In particular, the effect of *human impoliteness* on the perception of *robots' impoliteness* was not significant: an impolite robot, when treated impolitely by the human, was not perceived as less impolite, it was rather that its impoliteness was perceived as more fair or justifiable. That *human likeness* scores did not show statistically significant main effects of both independent variables, can apparently be interpreted that *human likeness* and *politeness* are perceived as two independent attributes.

Second, while the behavior of impolite robots, according to certain social attributes, was perceived *more positively* when the human was impolite, the perception was still *generally negative* compared to the behavior of polite robots. Therefore, the data shows that the participants generally considered it unacceptable for robots to be impolite – even in cases where humans were. Probably this attitude of the participants can explain the lack of a statistically significant interaction effect between the independent variables of "Human politeness" and "Robot politeness" according to scores for "*Socially acceptable*" and "*Friendly*".

Third, the data show a discrepancy between the "Robot's behavior" post-condition scale, which was targeting the robots' behavior directly, and the "Situation" scale, which was targeting the overall experience of the participants in the conditions of the study. We expected the data about the perception of the robots' behavior to be consistent with data about the perception of the overall situation in which the behavior takes place. However, we found that while an impolite robot *as such* was perceived more positively if the human was impolite, the assessment of the *whole situation* was reversed. The condition, in which both the robot and the human were impolite, was experienced most negatively, that is, as the most uncomfortable one. These data suggest that human experience of the social contexts that involve human-robot interactions is a complex, multifaceted phenomenon, and that the understanding of the contexts requires studying how humans experience other people, technologies, and the situation in which they encounter the people and the technologies.

## 5.3. Politeness reciprocity in human-robot interaction

The quantitative data, discussed in the previous section, are consistent with the second hypothesis, *H2: Human assessments of a robot's behavior are affected by an expectation of politeness reciprocity*. However, they do not directly support the hypothesis, as they only show a statistical interaction, not a causal relationship, between robot's and human (im)politeness.

Richer and more nuanced evidence regarding robot's politeness expectations, which complements and helps to interpret the quantitative data, comes from participants' free-text responses to "Concluding questions". Participants were asked to provide responses explaining the scores they assigned to the items comprising the scales of "Robot's likeability", "Empathy for robot", and "Empathy for other human". Many participants noted that they liked the robots better when they were polite towards polite humans than when they were polite towards impolite humans. The most common way of describing impolite robots responding to impolite humans was "not nice, but somewhat understandable" (see Table 7). The general notion of reciprocity, described as "fairness", was often mentioned as a reason for feeling empathy for a robot. Conversely, a lack of reciprocity in the scenario, in which a robot was rude to a polite human, was provided as a common reason for feeling empathy for the human.

The data generally confirm the second hypothesis (H2). For many of our participants, it was natural to expect politeness reciprocity and assess a robot's behavior by taking into consideration how the robot is treated by people. At the same time, the data show a wide range of individual perceptions, according to which the same robot's behavior can be assessed by different people as either perfectly acceptable or totally unacceptable. The data also reveal a diversity of reasons why it was considered acceptable or not acceptable to be rude to rude robots. These results are generally consistent with the diversity of opinions found in previous research (e.g., [10,27,33]).

An intriguing finding, which, in our view, warrants further investigation, is a difference between assessments based on concrete experiences and those based on abstract reflections. The "Concluding questions" section of the survey included two similar questions: how the participants liked a *particular robot* (*Red*), which was rude to a rude customer, and what they *generally thought* about robots being rude to rude humans. In their concrete assessments (i.e., answers to the first question), most participants were somewhat positive toward the robot, emphasizing that the robot's behavior, while not nice, was somewhat understandable. In cases of abstract reflections (i.e., answers to the second question), the participants were mostly negative, emphasizing that robots do not have feelings that would justify exhibiting politeness reciprocity.

The finding is, arguably, consistent with Takayama's [38] distinction between "in-the-moment" and "reflective" perception of interactive technological artifacts. According to Takayama, people dealing with such artifacts may perceive them "in-the-moment" as having their own agency, even though abstract reflections may clearly indicate that this is not the case. In the context of our study, an assessment of a *concrete* impolite robot (e.g., *Red*) could be similar to "in-the-moment" perception, so the robot was viewed as more of an agentic entity, and its behavior was perceived as more "understandable".

### 5.4. Limitations and prospects for future research

Several limitations of the study, reported in this paper, should be mentioned. First of all, the participants were essentially 'included observers' who were present in the situation but did not take initiative in the interactions with the robots. While, as argued in the 'Introduction' section, understanding the experience of this category of people is a relevant and important research topic, further studies are needed to investigate whether participants' perceptions would be different if they played a more active role in interactions with (im)polite robots.

There are also possibilities for exploring additional independent variables and experimental conditions when studying the perception of humans' and robots' mutual (im)politeness. First, the interaction scripts in our study were designed such that the human was the one starting to act either politely or impolitely, and the robot responded to the human's choice of interaction style. An alternative arrangement to explore would be to make a robot, rather than a human, set the tone of the interaction. Furthermore, the study only included polite or impolite conditions. Adding a neutral condition (or conditions) may provide new insights into how people assess robots' behavior and what types of interaction they prefer. Other promising directions for further research, in our view, would be exploring longer sequences of human-robot interaction and more complex patterns of politeness reciprocity, as well as analyzing how the social perception of robots is affected by the way they are treated by other robots. Finally, we found a selective effect of human politeness on the perception of some of the robots' behavior. While some attributes (*Justifiable, Competent, Responsive,* and *Fair*) showed a statistically significant interaction effect of the factors of human and robot politeness, others (*Socially acceptable, Friendly, Polite,* and *Human-like*) did not show such effect. Further studies are needed to find out the reasons why the effect of human politeness is selective.

## 6. Conclusion

HRI research has provided ample evidence that people commonly perceive robots as social actors and engage in social behavior toward them. However, there is less understanding of how robots should respond to such behavior. Our study, which specifically focused on politeness, found that the perception of how justifiable, competent, responsible, and fair a robot's behavior is depends on the interaction between human and robot's (im)politeness. Particularly, impolite robots received more positive assessments when humans behaved impolitely as well. It is important to note that we do not suggest designing social robots to interact impolitely with people. While some evidence suggests potential advantages of impolite robot behavior [15], our data reveal a clear preference for polite robots among participants. An overall conclusion from our study is more general, namely, that robots' social competence should include an ability to properly respond to human behavior directed at the robots themselves.

**Acknowledgements:** The research reported in this paper was supported by The Swedish Research Council, grant 2021-05409.

## Declarations

**Data availability statement:** The data that support the findings of this study are available upon reasonable request to the contact author, pursuant to data sharing-related policies at Umeå University.

**Conflict of interest:** The authors declare no conflict of interest.